\documentclass[12pt]{article}
\usepackage[utf8]{inputenc}
\usepackage{lineno}          
\usepackage{authblk}         
\usepackage{geometry}        
\usepackage{hyperref}
\usepackage{amsmath}  
\usepackage{setspace}        
\usepackage{multirow}
\usepackage{longtable}
\usepackage{multicol}
\usepackage{graphicx}
\usepackage{placeins}
\usepackage{algorithm}
\usepackage{algorithmic}
\usepackage[labelsep=space]{caption}
\usepackage[numbers]{natbib} 

\title{When Fireflies Cluster; Enhancing Automatic
	Clustering via Centroid-Guided Firefly
	Optimization}
\author[1]{MKA Ariyaratne \thanks{Corresponding author: mkanuradha@sjp.ac.lk}}
\author[2]{Azwirman Gusrialdi}
\author[3]{Yury Nikulin}
\author[2]{Jaakko Peltonen}

\affil[1]{Department of Computer Science, Faculty of Applied Sciences, University of Sri Jayewardenepura, Sri Lanka}
\affil[2]{Faculty of Engineering and Natural Sciences,Tampere University, Tampere, Pirkanmaa, Finland}
\affil[3]{Department of Mathematics and Statistics, University of Turku, Finland}
\affil[4]{Faculty of Information Technology and Communications, Tampere University, Tampere, Pirkanmaa, Finland}
\date{}

\begin{document}
	
	\maketitle
	\begin{abstract}
		This work presents a novel variant of the Firefly Algorithm (FA) for data clustering, addressing limitations of traditional methods like K-Means that struggle with non-uniform cluster shapes, densities, and the need for pre-defining the number of clusters. The proposed algorithm introduces a centroid movement strategy and a multi-objective fitness function that balances compactness, separation, and a novel TSP-based navigation penalty. It automatically estimates the optimal number of clusters and dynamically adjusts cluster boundaries. Application to robotic sensor networks highlights its practical value, with experiments showing improved clustering quality and reduced intra-cluster path distances compared to K-Means. These results confirm the algorithm’s robustness in complex spatial clustering tasks, with potential for future extensions to higher-dimensional and adaptive scenarios.
	\end{abstract}
	
	\setstretch{1.1}
	\section{Introduction}
	Clustering is the process of categorizing a given dataset into homogeneous groups based on their shared characteristics, where similar data points are grouped together and dissimilar ones are placed in separate groups. It stands as a crucial unsupervised learning task, aiming to uncover patterns within unlabeled data. Given its broad applicability, cluster analysis has emerged as a prominent research area in data mining, finding utility across various scientific disciplines. Despite the proliferation of clustering algorithms in recent years and their widespread adoption in fields like computational biology, mobile communication, medicine, economics, and more, standardization remains a challenge. One of the primary issues with clustering algorithms is their lack of universal applicability; an algorithm may perform well with one type of dataset but yield subpar results with another. Despite efforts to develop standardized algorithms capable of performing effectively across all scenarios, significant progress in this regard remains elusive. Numerous clustering algorithms have been proposed, each with its own strengths and weaknesses, making it challenging to find a one-size-fits-all solution for real-world scenarios \cite{jain1999data}.\\
	
	\noindent
	A limited number of clustering problems demonstrate polynomial boundedness. These algorithms share the property that a specific approach ensures the number of computational steps remains within a polynomial function of the input length \cite{10.1007/978-3-642-95322-4_5}. On the other hand, many clustering problems encountered in practical applications are classified as NP-hard, implying that their computational complexity grows exponentially as the input size increases \cite{welch1982algorithmic}.\\
	
	\noindent
	Clustering comes into play when there's no prior knowledge about how data points should be classified into specific groups. Because of this, it's generally considered a form of unsupervised learning. However, standard clustering techniques—such as K-means \cite{ahmed2020k}, fuzzy c-means \cite{suganya2012fuzzy}, K-medoids \cite{park2009simple}, x-means \cite{pelleg2000x}, and the Nelder-Mead simplex search \cite{singer2009nelder}—come with notable drawbacks. They often rely heavily on initial parameter settings, struggle to avoid local optima, and don't scale well for larger datasets. As a result, these conventional methods are often inadequate for handling large-scale clustering tasks effectively.\\
	
	\noindent
	Optimization is a common need across various domains like science, engineering, economics, computing, and biology. Many times, these optimization tasks involve managing multiple constraints and dealing with large-scale problems. In real-world scenarios, traditional optimization methods, like gradient-based techniques, often struggle due to inherent limitations. Issues such as reliance on initial guesses and the requirement for continuous, differentiable problem representations make these methods less practical for solving real-world problems \cite{boyd2004convex}.\\
	
	\noindent
	It can be stated that current state of the art solution for addressing these challenges involves the utilization of meta-heuristics, particularly algorithms inspired by nature. These approaches are gaining popularity due to their robustness, flexibility, and effectiveness in tackling practical optimization problems. Nature-inspired algorithms generally fall into two main categories. The first includes evolutionary computing methods, such as Genetic Algorithms (GA) \cite{holland1992adaptation}  and Differential Evolution (DE) \cite{storn1997differential}, which focus on mimicking natural selection and adaptation. The second category consists of swarm intelligence techniques like Particle Swarm Optimization (PSO) \cite{kennedy1995particle}, Ant Colony Algorithms (ACO) \cite{dorigo1999ant}, Firefly Algorithm \cite{yang2009firefly}, and Cuckoo Search Algorithm (CA) \cite{yang2009cuckoo}, among others, which draw inspiration from collective behaviors observed in nature. While swarm intelligence algorithms have gained significant traction over the past five decades, many still require fine-tuning to maximize their efficiency and overcome inherent challenges.
	Hence, selecting the most suitable algorithm for a given problem has become a crucial aspect when employing swarm intelligence algorithms.\\
	
	\noindent
	The necessity for efficient and accurate clustering algorithms tailored for big data has become increasingly apparent. Due to the complexity of the problem, recent research has shown a tendency towards utilizing meta-heuristic approaches, such as nature-inspired algorithms, to address these challenges. Meta-heuristics like Tabu search, simulated annealing, GA, DE, FA, PAO, and Glow Warm Search Algorithm (GWA) have been explored for clustering tasks, reflecting efforts to overcome the limitations of conventional methods \cite{al1995tabu, maulik2000genetic, kuila2014novel, ariyaratne2022comprehensive, ahmadi2021modified, kuo2011application}.
	
	\subsection{Research Question}
	 The study done by Made Widhi Surya Atman et. al. examines deployment of robotic sensor networks for persistent and effective monitoring of multiple locations of interest across a large, grid-represented field \cite{10379586}.
	To achieve this, they have proposed a novel two-layer control framework. The first layer, referred to as the task allocation strategy, involves organizing the targeted grid regions into smaller, manageable tasks (i.e., regions of interest). Robots are then optimally assigned to these tasks based on their initial positions (e.g., the locations of their base stations) and sensing capabilities.
	
	The targeted research \cite{10379586} primarily used K-Means for clustering; however, the study encountered challenges related to several of these factors, highlighting the need for more adaptive and robust clustering techniques.
	\begin{enumerate}
		\item need an algorithm that does not depend on initial cluster center selection.
		\item need an algorithm, that gives clusters without specifying exact K value (providing a range for K)
		\item The clustering objective does not solely depend on the proximity; they need to cluster with efficient navigation within clusters.  
	\end{enumerate}
	Through this study we tried to provide meaningful answers to these problems by utilizing firefly algorithms. We mainly consider solving the problem of effective clustering by implement a variant of the firefly algorithm which is capable of clustering
	\begin{enumerate}
		\item That does not depend on the initial cluster centers.
		\item That automatically chooses the best number of clusters K for the problem at hand.
		\item That considers the \textbf{efficient navigation} within clusters, when selecting cluster centers.
	\end{enumerate}

	For the convenience of the reader, the paper is structured as follows. Section 2 provides a comprehensive review of the existing literature on data clustering using meta-heuristic approaches. It focuses on key aspects such as the management of initial cluster centers, determination of the number of clusters, and dynamic clustering, while highlighting the influence of prior studies on the present research. Section 3 presents the materials and methods used in the study. It introduces the Firefly Algorithm, discusses its biologically inspired foundations, and explains how meta-heuristic algorithms are adapted for clustering tasks. The section also details the representation of the population, fitness functions, and proposed mechanisms for determining the number of clusters and handling dynamic clustering. Section 4 reports and discusses the results obtained from the experiments, evaluating the proposed algorithm through a comparative analysis with the traditional K-Means algorithm. Finally, Section 5 concludes the paper by summarizing key findings and offering recommendations for future research directions.
	
	\section{Clustering with Intelligence: Surveying Meta-Heuristic Strategies}
	
	Cluster Analysis is a topic of interest for many decades. There are enormous algorithms and related work on the topic and therefore it is not and easy task to put them together. Many review articles can be found categorizing  different approaches of clustering.  In this related work section, we review clustering research conducted using nature-inspired optimization algorithms. Here, we try to summarize most of such work as common and in later sections specifically the contribution of FA for clustering in different environments.
	Clustering problems are generally considered NP-hard (finding the optimal partitioning of a dataset into clusters often requires searching through an
	exponentially large number of possibilities). Therefore one can claim that meta-heuristic approaches are particularly
	effective for solving clustering problems because they are well-suited for tackling NP-hard optimization problems.
	They have following benefits over exact methods such as Exhaustive Enumeration (Brute Force Search), Branch and Bound and others.
	\begin{itemize}
		\item Overcoming Local Optima
		\item Flexibility with Objective Functions
		\item Handling Arbitrary Cluster Shapes
		\item Robustness to Initialization
		\item Handling Mixed and High-Dimensional Data
		\item Scalability and Parallelism
	\end{itemize}

Therefore, clustering with meta-heuristics / nature inspired algorithms, has a rich literature. Several studies have demonstrated the effectiveness of meta-heuristic algorithms in addressing clustering challenges across various domains. Akay et al. introduced a genetic algorithm (GA) enhanced with an innovative fitness function specifically designed for clustering tasks \cite{akay2020genetic}. Their approach involved refining chromosome structures and generating new populations using genetic operators. When evaluated on synthetic datasets, the proposed GA outperformed traditional clustering methods such as K-means and Ward's method in terms of accuracy and robustness \cite{hartigan1979algorithm, blashfield1980growth}. In a different application domain, Behravan et al. proposed an automatic Particle Swarm Optimization (PSO) clustering algorithm to analyze football datasets, aiming to identify distinct player roles during matches \cite{behravan2019finding}. Their big data clustering method leveraged swarm intelligence to determine optimal cluster centroids and was shown to outperform traditional clustering techniques across six synthetic datasets. Document clustering, another key area of research, involves grouping textual documents based on content similarity and is vital for applications such as topic extraction, text mining, and information retrieval \cite{karol2013evaluation}. Due to the limitations of classical algorithms like K-means, hybrid approaches combining PSO with K-means and Fuzzy C-Means have been proposed to improve clustering accuracy \cite{karol2013evaluation, cui2005document}. These nature-inspired algorithms have consistently demonstrated enhanced precision in document grouping. Similarly, meta-heuristic techniques such as PSO, the bees algorithm, the firefly algorithm (FA), and genetic algorithms have been effectively applied to image clustering tasks, addressing complex pattern recognition problems and yielding improved clustering performance \cite{omran2005particle, wong2011image, hancer2012artificial, hrosik2019brain, scheunders1997genetic}.

\subsection{The Role of FA in Data Clustering}

Utilizing algorithms inspired by nature for data clustering has long been explored in optimization literature, dating back a decade or two. The field has seen numerous adaptations of firefly algorithms to tackle clustering challenges. Here the consideration goes with existing body of research employing the firefly algorithm for diverse clustering tasks. While algorithms may vary in their presentation, organizing descriptions into categories based on how the firefly algorithm has been integrated with other techniques to address clustering problems will facilitate later discussions. A summary of the review is provided in Table  \ref{tab11}.
\FloatBarrier
\renewcommand{\arraystretch}{1.5}
\footnotesize

\begin{longtable}{|p{5cm}|c|l|p{5cm}|}
	\caption{Detailed Summary of Research Papers on Firefly Algorithm (FA) for Data Clustering} \label{tab11} \\
	
	\hline
	\textbf{Paper Name} & \textbf{Year} & \textbf{FA Variant} & \textbf{Summary and Performance} \\ \hline
	\endfirsthead
	
	\hline
	\textbf{Paper Name} & \textbf{Year} & \textbf{FA Variant} & \textbf{Summary and Performance} \\ \hline
	\endhead
	
	\hline
	\endfoot
	
	\hline
	\endlastfoot
	
	Clustering using firefly algorithm: performance study \cite{SENTHILNATH2011164} & 2011 & Standard FA & Solved 13 benchmark problems; average error percentages were lower than other nature-inspired algorithms. \\ \hline
	A new hybrid approach for data clustering using firefly algorithm and K-means \cite{hassanzadeh2012new} & 2012 & Hybrid FA + K-means & Overcame initial sensibility of K-means by using FA to find optimal centroids, reducing intra-cluster distance. \\ \hline
	Performance analysis of firefly algorithm for data clustering \cite{banati2013performance} & 2013 & FClust (FA+PSO-like) & Applied to web information retrieval; demonstrated faster convergence and higher stability than PSO and DE. \\ \hline
	Implementation of energy efficient clustering using firefly algorithm in wireless sensor networks \cite{sarma2014implementation}& 2014 & FA & Optimized energy consumption in WSN; prolonged network lifetime compared to PSO and LEACH. \\ \hline
	A hybrid firefly algorithm with fuzzy-C mean algorithm for MRI brain segmentation \cite{alsmadi2014hybrid} & 2014 & FAFCM (FA + FCM) & Segmented brain tumors without prior info while avoiding the hereditary deficiencies of FCM. \\ \hline
	An improved firefly fuzzy c-means (FAFCM) algorithm for clustering real world data sets \cite{nayak2014improved}& 2014 & Improved FAFCM & Used FCM objective function to evaluate fireflies; provided faster convergence and optimum values. \\ \hline
	Segmentation of MRI Brain Images Using FCM Improved by Firefly Algorithms \cite{Alarticle}& 2014 & FA-based Fuzzy Clustering & Two-phased approach; FA obtained near-optimum centers to improve FCM efficiency significantly. \\ \hline
	A hybrid data clustering using firefly algorithm based improved genetic algorithm \cite{kaushik2015hybrid} & 2015 & FA + Genetic Algorithm & FA boosted the GA initialization process, resulting in improved inner and intra-cluster distances. \\ \hline
	Synchronous firefly algorithm for cluster head selection in WSN \cite{baskaran2015synchronous}  & 2015 & Synchronous FA & Combined FA with GA crossover/mutation; minimized packet loss and energy consumption in WSN. \\ \hline
	Firefly algorithm based clustering technique for Wireless Sensor Networks \cite{manshahia2016firefly}& 2016 & Canonical FA & Selected cluster heads based on energy and distance; achieved higher packet delivery ratio than Energy-Aware methods. \\ \hline
	Image segmentation using firefly algorithm \cite{sharma2016} & 2016 & KFA (FA + K-means) & Image segmentation; obtained higher correlation coefficients, indicating better quality segmentation. \\ \hline
	Taiwanese export trade forecasting using firefly algorithm based K-means algorithm and SVR with wavelet transform \cite{kuo2016taiwanese}& 2016 & FA + K-means + SVR & Multi-stage model for export trade values; outperformed existing forecasting algorithms. \\ \hline
	Fireflies can find groups for data clustering \cite{mizuno2016fireflies}& 2016 & Hybrid FK (FA + K-means) & Successfully identified suitable clusters to serve as initial seeds for K-means. \\ \hline
	Data Clustering Using Improved Fire Fly Algorithm \cite{sadeghzadeh2016data}& 2016 & Hybrid FA + DE & Split population into best (FA) and worst (DE) parts; outperformed standard FA and K-means. \\ \hline
	Protein complex identification through Markov clustering with firefly algorithm on dynamic protein-protein interaction networks \cite{lei2016protein}& 2016 & FA + Markov Clustering & FA optimized the inflation parameter of MCL; superior performance in identifying protein complexes. \\ \hline
	Improvement of the Firefly-based K-means Clustering Algorithm \cite{zhou2018improvement} & 2018 & PFK / GPFK & Introduced Probabilistic (PFK) and Greedy Probabilistic (GPFK) variants to reduce complexity. \\ \hline
	Chaotic firefly algorithm-based fuzzy C-means algorithm for segmentation of brain tissues in magnetic resonance images \cite{ghosh2018chaotic}& 2018 & C-FAFCM (Chaotic FA) & Used chaotic maps for parameter tuning; increased global search accuracy and execution speed. \\ \hline
	Data aggregation in wireless sensor networks using firefly algorithm \cite{mosavvar2019data}& 2019 & FA & Reduced duplicate data in WSN; more efficient than LEACH and SFA models. \\ \hline
	Improving K-means clustering with enhanced firefly algorithms \cite{xie2019improving}& 2019 & IIEFA / CIEFA & Used randomized control matrices and scattering to increase search diversity and efficiency. \\ \hline
	Automatic data clustering using hybrid firefly particle swarm optimization algorithm \cite{agbaje2019automatic}& 2019 & FAPSO (Hybrid FA-PSO) & Automatically determined the number of clusters (K) using CS and DB indices. \\ \hline
	A hybrid firefly algorithm with particle swarm optimization for energy efficient optimal cluster head selection in wireless sensor networks \cite{pitchaimanickam2020hybrid}& 2020 & HFAPSO (Hybrid FA-PSO) & Improved network lifetime, residual power, and throughput by incorporating Pbest and Gbest values. \\ \hline
	Improved density peaks clustering based on firefly algorithm \cite{zhao2020improved}& 2020 & IDPC-FA & Balanced kernels with weight factors to improve cluster performance over standard DPC. \\ \hline
	Combined fuzzy clustering and firefly algorithm for privacy preserving in social networks \cite{langari2020combined}& 2020 & KFCFA (Fuzzy + FA) & Optimized K-member clusters to ensure data anonymity in social network graphs. \\ \hline
	
\end{longtable}
\FloatBarrier
\normalsize
	
	\section{Preliminaries}
	The initial step in deploying the robotic sensor network involved developing a strategy to filter out unimportant areas by dividing the domain into smaller regions, with each region representing a cluster of critical locations requiring continuous monitoring. In \cite{atman2024two}, K-Means clustering was employed for this task; however, the method exhibited certain limitations. One of their tasks is to find a method to cluster the data automatically without specifying the K value (number of clusters). To achieve the task, the original firefly algorithm was used with slight modification to the concept of movements in the algorithm.
	
	We have implemented two versions of the algorithm, one with specific K value (K is given) and the other one without giving a specific K value (Instead, a range for the K values is provided).  The chapter introduces the framework of the algorithms. The frame works of the algorithms are provided in Algorithm Pseudo-codes \ref{al:algorithm1}, \ref{al:algorithm2}, \ref{al:algorithm3}, \ref{al:algorithm4}, \ref{al:algorithm5}. 

\subsection{Materials - Preliminaries}
\subsubsection{Firefly Algorithm}
Literature points out many meta-heuristic algorithms, that has been tried for the problem of clustering. Given the NP-hard nature of clustering problems, meta-heuristic approaches have been widely explored in research to achieve improved clustering solutions. A key challenge in this domain is selecting appropriate algorithms while ensuring effective evaluation of cluster quality—an area that remains an open topic for further investigation.

The firefly, commonly known as the lightning bug, emits light to attract mates or prey. Inspired by their flashing behavior, Xin-She Yang introduced the Firefly Algorithm (FA) in 2009 \cite{yang2009firefly}. The algorithm is based on the following principles:
\begin{itemize}
	\item Firefly attraction is gender-independent.
	\item Fireflies move toward brighter ones, with brightness decreasing over distance. If no brighter firefly is found, movement is random.
	\item Brightness is determined by the objective function specific to the problem.
\end{itemize}

Biological insights into fireflies' behavior have played a crucial role in shaping this optimization technique \cite{EVO:EVO1199}.

Like most meta-heuristics, the FA begins with a randomly generated initial population. Its parameters must be properly set before execution. Once initialized, fireflies move toward brighter ones based on the following equation.

\begin{equation}
	x_i=x_i+\beta(x_j-x_i )+\alpha(rand-0.5)
	\label{eqn:move}
\end{equation} \\
where
\begin{equation}
	\beta=\beta_0. e^{-\gamma r^2}
	\label{eqn:attraction}
\end{equation} \\
\noindent
In this context, $x_i$ and $x_j$ represent two distinct fireflies. The parameter $\beta$ quantifies the degree of attraction between them, while $\alpha$ governs the step size of their movement. The attraction strength at $r = 0$ is denoted by $\beta_0$, where $r$ signifies the spatial separation between two fireflies.

The three components in \texttt{equation \ref{eqn:move}} contribute to different aspects of firefly behavior: the first term accounts for the firefly's current position, the second reflects its attraction toward another firefly, and the third introduces a stochastic factor. This equation effectively balances both exploration and exploitation, which are crucial for achieving an optimal trade-off between global and local search strategies.

The parameter $\alpha$ plays a pivotal role in the randomization mechanism, which follows either a Uniform or Gaussian distribution. To regulate this randomness, Yang introduced a randomness reduction factor, $\delta$, which progressively decreases $\alpha$ after each iteration, following the adjustment rule described in \texttt{equation \ref{eqn:reduction}}.

\begin{equation}
	\alpha=\alpha.\delta \quad
	where \quad \delta\in [0,1]
	\label{eqn:reduction}
\end{equation}

FA, a well-established meta-heuristic, has proven highly effective in handling optimization problems. Its applications span diverse fields, with thousands of studies utilizing FA in areas such as clustering \cite{SENTHILNATH2011164}, image processing \cite{KANIMOZHI20151099, RAJINIKANTH20151449}, and forecasting \cite{XIAO2016135, IBRAHIM2017413}. Various FA variants have been introduced, including discrete adaptations \cite{DBLP:journals/corr/OsabaYDOMP16, xxx}, hybrid models \cite{articleHybrid, XIA2018488, SAHU20159}, and multi-objective approaches \cite{Yang2013x, ZHAO2017549, LV2018}.

Comprehensive reviews on FA and its applications are available in \cite{fister2013comprehensive, Tilahun2017, Tilahun20172}. In 2013, Yang et al. introduced a self-tuning algorithm framework, successfully implemented with FA \cite{yang2013framework}. The pseudocode of the original FA is provided in Algorithm \ref{al:algorithm1} \cite{yang2018firefly}.

This research highlights the superior clustering capabilities of the firefly algorithm, attributing its effectiveness to several key factors:
\begin{enumerate}
	\item The firefly algorithm demonstrates greater promise compared to other meta-heuristics, such as particle swarm optimization, particularly in handling multi-modal functions with enhanced naturalness and efficiency.
	\item Notably, particle swarm optimization and several other meta-heuristics can be regarded as specialized instances of the firefly algorithm, further underlining its versatility and adaptability.
	\item The structure of the firefly algorithm naturally enables the automatic division of the entire swarm into several sub-swarms. This occurs because the attraction between nearby fireflies is stronger than that between distant ones, and the swarm’s division is influenced by the average range of attractiveness variations.
\end{enumerate}

\FloatBarrier
\begin{algorithm}[!h]
	\footnotesize
	\setstretch{0.9}
	\caption{: Pseudocode of the FA as given in \cite{yang2018firefly}}
	\label{al:algorithm1}
	\begin{algorithmic}[1]
		\STATE Start
		\STATE Set initial parameter values ($\alpha$, $\beta_0$, $\delta$, and $\gamma$) based on the problem requirements.
		\STATE Define the objective function $f(X)$.
		\STATE Generate an initial population of fireflies, $X_i$ for $i = 1, 2, ..., n$.
		\STATE Compute the fitness $I_i$ of each firefly $X_i$ using $f(X_i)$.\\[0.5cm]
		
		\WHILE{Termination condition is not met}  
		\FOR{$ i = 1$ to $n$ (iterate through all fireflies)}  
		\FOR{$ j = 1$ to $n$ (compare with all other fireflies)}  
		\IF{$ I_j > I_i $}  
		\STATE Move firefly $i$ toward firefly $j$ using equation (\ref{eqn:move}).  
		\ENDIF  
		\STATE Adjust attractiveness based on distance $r$ using the function $e^{-\gamma r^2}$ as defined in equation (\ref{eqn:attraction}).  
		\STATE Evaluate new solutions and update light intensity accordingly.  
		\ENDFOR  
		\ENDFOR  
		\STATE Rank fireflies and determine the current best solution.  
		\ENDWHILE  
		\STATE Process the final results and visualize the outcome.  
		\STATE End
	\end{algorithmic}  
\end{algorithm}
\FloatBarrier

In addition to its inherent advantages, the firefly algorithm boasts ease of implementation, efficiency, adaptability, and low computational overhead compared to alternative meta-heuristics. Furthermore, it excels in tackling complex and discrete optimization challenges, consistently delivering near-optimal solutions.

\subsection{Methods: Adapting Algorithms to solve clustering problems}
The clustering problem addressed in most of the studies can be formally defined as follows.

Given a dataset:  
\[
X = \{x_1, x_2, ..., x_n\}
\]
where each data point \( x_i \) (\( i = 1,2,...,n \)) exists in a \( D \)-dimensional space, the goal is to partition \( X \) into \( K \) **non-overlapping clusters**:
\[
P = \{p_1, p_2, ..., p_K\}
\]
where each cluster \( p_i \) is represented by a centroid \( c_i \) (\( i = 1,2,...,K \)). These centroids serve as the representative points of their respective clusters.

The clustering process must satisfy the following conditions:

1. Clusters are mutually exclusive (no data point belongs to more than one cluster):
\[
p_i \cap p_j = \emptyset, \quad \forall i, j \in \{1,2,\dots,K\}, \quad i \neq j
\]

2. Clusters together cover the entire dataset:
\[
p_1 \cup p_2 \cup \dots \cup p_K = X
\]

3. Each cluster is a non-empty subset of \( X \):
\[
p_i \subseteq X, \quad p_i \neq \emptyset, \quad \forall i \in \{1,2,\dots,K\}
\]

These conditions ensure that each data point is assigned to exactly one cluster and that all clusters are well-defined subsets of \( X \).

\subsubsection{Firefly algorithm based clustering}
Use of meta-heuristic algorithm to solve a given problem has two main tasks. Apart from those mandatory tasks, in our research, we introduced a new movement strategy, as our fireflies are of varying lengths.
\begin{enumerate}
	\item Find a suitable encoding mechanism to represent solutions
	\item Formulation of appropriate fitness function/s
	\item Define \textbf{movement function/formula} for fireflies with varying lengths.
\end{enumerate}

\subsubsection{Encoding Method}

In this thesis, a firefly will represent set of cluster centroids. Therefore each solution appears as a multi dimensional array of $K$x$D$ elements. $K$ represent the number of clusters and $D$ represent the dimension of data. For an example; a situation of $K = 3$ and $D = 2$, with 5 fireflies, a firefly/solution can be represented as shown in the \texttt{Table \ref{Tab3.1}}.\\
\FloatBarrier

\begin{table}[!h]
	\centering
	\begin{tabular}{|ll|l|l|l|}
		\hline
		\multicolumn{2}{|l|}{\textbf{K}}                               & \textbf{1} & \textbf{2} & \textbf{3} \\ \hline
		\multicolumn{1}{|l|}{\multirow{2}{*}{\textbf{D}}} & \textbf{1} & 2.6        & 7.1        & 12.7       \\ \cline{2-5} 
		\multicolumn{1}{|l|}{}                            & \textbf{2} & 3.3        & 5.8         & 1.4        \\ \hline
	\end{tabular}
	\caption{Example representation of a solution}
	\label{Tab3.1}
\end{table}
\FloatBarrier
Here [2.6, 3.3], [7.1, 5.8], [12.7, 1.4] appear to be three centroids given in a particular iteration. With standard notations, this can be presented as given in \texttt{Table \ref{Tab 3.2}}. In this Table $C_{ij}$ is the
$j_{th}$ dimension of the $i_{th}$ centroid.
\FloatBarrier
\begin{table}[!h]
	\centering
	\begin{tabular}{|l|l|l|l|l|l|}
		\hline
		$C_{11}$ & $C_{12}$ & $C_{21}$ & $\bullet \quad \bullet \quad \bullet $  & $C_{K1}$ & $C_{K2}$\\ \hline
	\end{tabular}
	\caption{A solution comprising $K$ centroids for a two-dimensional dataset}
	\label{Tab 3.2}
\end{table}
\FloatBarrier
Initial population was formed selecting data points randomly to represent initial cluster centroids.

\subsubsection{Fitness Formula}

Meta-heuristic algorithms are often optimization algorithms, hence the clustering problem is converted into an optimization problem. The fitness function of the problem therefore has to evaluate the goodness of the clusters. 
In the literature, there can be found many evaluation methods used, to evaluate the goodness of the clusters. Many such methods consider the properties of inter cluster distance and the intra cluaster distance. But our objectives are different.Therefrore, to solve the proposed problem, some modifications were considered.
For an example, after clustering, we have to ensure that clustered data are navigation friendly. To fulfill this task, one option is to incorporate route finding optimization algorithm with the cluster formulation, but it is computationally expensive. Considering those facts,
we designed a fitness function according to the following
formula.

\begin{equation}
	\text{fitness} = \alpha \cdot \frac{\text{compactness}}{\text{max\_compactness}} + 
	\beta \cdot \frac{\text{separation}}{\text{max\_separation}} + 
	\gamma \cdot \frac{\text{total\_TSP\_penalty}}{\text{max\_TSP\_penalty}}
	\label{eqFitnessNormalized}
\end{equation}

Where:

\begin{itemize}
	\item \(\alpha\), \(\beta\), and \(\gamma\) are weights for normalized compactness, separation, and the TSP penalty, respectively.
	\item \(\text{max\_compactness}\), \(\text{max\_separation}\), and \(\text{max\_TSP\_penalty}\) are the maximum possible values for compactness, separation, and the TSP penalty, ensuring all metrics are normalized.
\end{itemize}
\begin{enumerate}
	\item Compactness\\
	Compactness measures the average distance between each point in a cluster and the centroid of that cluster. After normalization:
	
	\[
	\text{normalized\_compactness} = \frac{\text{compactness}}{\text{max\_compactness}}
	\]
	
	Where:
	
	\[
	\text{compactness} = \frac{1}{K} \sum_{k=1}^{K} \frac{1}{|C_k|} \sum_{x \in C_k} \| x - \mu_k \|
	\]
	
	\begin{itemize}
		\item \(K\) is the number of clusters,
		\item \(C_k\) is the set of points in the \(k\)-th cluster,
		\item \(\mu_k\) is the centroid of the \(k\)-th cluster,
		\item \(\| x - \mu_k \|\) is the Euclidean distance between a point \(x\) and the centroid \(\mu_k\),
		\item \(\text{max\_compactness}\) is the maximum compactness value, determined empirically or through a theoretical bound.
	\end{itemize}
	
\item Separation\\
Separation measures the average distance between centroids of different clusters. After normalization:

\[
\text{normalized\_separation} = \frac{\text{separation}}{\text{max\_separation}}
\]

Where:

\[
\text{separation} = \frac{1}{K(K-1)} \sum_{k=1}^{K} \sum_{l=k+1}^{K} \| \mu_k - \mu_l \|
\]

\begin{itemize}
	\item \(\mu_k\) and \(\mu_l\) are centroids of clusters \(k\) and \(l\), respectively,
	\item \(\| \mu_k - \mu_l \|\) is the Euclidean distance between the two centroids,
	\item \(\text{max\_separation}\) is the maximum separation value, determined empirically or through a theoretical bound.
\end{itemize}

\item Total TSP Penalty\\
The total TSP penalty is calculated based on the traveling salesman problem (TSP) distance within each cluster. After normalization:

\[
\text{normalized\_TSP\_penalty} = \frac{\text{total\_TSP\_penalty}}{\text{max\_TSP\_penalty}}
\]

Where:

\[
\text{total\_TSP\_penalty} = \sum_{k=1}^{K} \frac{\text{TSP\_distance}_k}{\log(1 + |C_k|)}
\]

\begin{itemize}
	\item \(\text{TSP\_distance}_k\) is the TSP distance for cluster \(k\),
	\item \(|C_k|\) is the number of points in cluster \(k\),
	\item \(\text{max\_TSP\_penalty}\) is the maximum TSP penalty value, determined empirically or through a theoretical bound.
\end{itemize}

The firefly who has a lesser fitness value is considered to be more fit among the fireflies.

The specialty of the above fitness function lies in its multi-objective approach to optimize the quality of the clustering solution. By normalizing compactness, separation, and TSP penalty, the formula ensures that all components contribute equally to the overall fitness evaluation, regardless of their scales. This approach provides a balanced assessment of cluster quality.
Here's what makes this fitness function unique:
\begin{enumerate}
	\item Multi-Objective Design: \\
	The function simultaneously optimizes three objectives:
	\begin{itemize}
		\item Compactness: Encourages tight clusters where data points are close to their respective centroids, thus improving within-cluster similarity.
		\item Separation: Promotes distinct, well-separated clusters by maximizing the distance between cluster centroids, ensuring clusters are distinct and far from each other.
		\item TSP Penalty (Traveling Salesman Problem Penalty): A penalty that enforces a spatial organization of points within each cluster, encouraging smooth transitions between data points within clusters. This is particularly important for applications where path optimization or spatial arrangement is important, like routing or logistics.
	\end{itemize}
	\item Penalty for Small Clusters:\\
	The TSP penalty term introduces a penalization mechanism for small clusters. By dividing the TSP distance by 
	$log(1+|C_k|)$, the function ensures that small clusters (with fewer data points) receive a higher penalty. This discourages forming very small clusters that may be less meaningful or harder to generalize.
	
	\item Weighted Objective Balance:\\
	The inclusion of three weight parameters $\alpha, \beta, \gamma$ gives flexibility to balance the importance of each objective (compactness, separation, and TSP penalty). These weights can be tuned based on the specific problem requirements, allowing the fitness function to prioritize certain objectives over others depending on the context.
	
	\item Incorporating TSP for Intracluster Quality:\\
	The use of the Traveling Salesman Problem (TSP) distance within each cluster is uncommon in typical clustering fitness functions. This term ensures that points within a cluster are not just close to the centroid but are arranged in a spatially optimal way, which can be important for spatial or route-based clustering problems. This is particularly useful in cases where sequentially visiting the data points within a cluster is required, such as in logistics, sensor networks, or route planning.
	
	\item Generalization to Different Domains:\\
	The fitness function is flexible and general enough to be applied in a wide range of domains where clustering is required, but the inclusion of spatial and routing constraints (through the TSP penalty) makes it particularly valuable in applications such as geospatial clustering, transportation, and supply chain management, where distances and routes between data points are critical.

\end{enumerate}

\end{enumerate}

In summary, the key specialty of this fitness function is its comprehensive assessment of clustering quality, which balances internal cohesion (compactness), external separation, and a spatial/routing consideration via the TSP penalty. This makes it especially useful in applications where spatial relationships and path optimization are critical.

\subsubsection{Clustering with known K values}
Clustering with known K values is the most common and the direct approach to clustering. However, we employed firefly algorithm to cluster data with known K values, in order to see its initial performances. The canonical FA is used, without any modifications. The proposed problem specific fitness function was used (equation \ref{eqFitnessNormalized}).

\subsubsection{Clustering with unknown K values - Proposed MODIFIED FIREFLY ALGORITHM}
Clustering with unknown K values or automatic clustering is an interesting topic in clustering analysis. Most of the approaches in literature run the algorithms many times with different K values (similar to the  Elbow Method in K means clustering) and based on the results, select the most suitable K value. There were some meta-heuristic approaches that used flags (active/inactive) states to represent number of clusters as well \cite{das2007automatic}. 

We propose a method of passing K as a parameter to the algorithm with a feasible range, and the K with highest fit centroids are taken as the suitable K value and the centroids. 

The following steps were performed to implement the firefly algorithm to solve the clustering problem with automatic selection of K values.

\begin{enumerate}
	\item Generate initial fireflies (initial centroids)
	\begin{enumerate}
		\item For each firefly
		\begin{enumerate}
			\item Generate a random number of clusters between K\_min and K\_max (say K).
			\item Take a random sample of size K from the data set.
		\end{enumerate}
		
	\end{enumerate}
	\item Calculate fitness of each firefly \textbf{(equation \ref{eqFitnessNormalized})}
	
	\item Move fireflies towards more fitter fireflies (consider minimization (equation \ref{moveFunc})).
	\begin{enumerate}
		\item This function needs to be modified as the length of the fireflies (how many centroids a firefly has) depends of its K value and it is not same for all, in this implementation.
		\begin{enumerate}
			\item Compare two fireflies in the current iteration.
			\item Take the centroids of each firefly 
			\item If \textbf{Firefly\_2} is fitter than \textbf{Firefly\_1}, Move centroids of \textbf{Firefly\_1} towards the nearest centroids in \textbf{Firefly\_2}.
			\item When moving a centroid, use the standard movement function in the canonical firefly algorithm.
			\item After moving centroids, towards the nearest centroids of the fitter firefly, Probabilistically adjust the number of centroids in the \textbf{Firefly\_1}.
		\end{enumerate}
	\end{enumerate}
	\item Find suitable centroids from the data set, for the moved new fireflies (new centroid sets), as the moved centroids are not in the data set.
	\item Evaluate the new fireflies (use \textbf{(equation \ref{eqFitnessNormalized})}).
	\item Repeat the process until the end criteria meets.
\end{enumerate}

The pseudo codes for the above steps are given in Algorithm \ref{al:algorithm2},  Algorithm \ref{al:algorithm3},  Algorithm \ref{al:algorithm4} and, Algorithm \ref{al:algorithm5}.

\FloatBarrier
\begin{algorithm}[!h]
	\footnotesize
	\caption{Main Function-Automatic Clustering}
	\label{al:algorithm2}
	\begin{algorithmic}[1]
		\STATE Initialize Firefly Parameters: $\alpha = 0.5$, $\gamma = 1$, $\delta = 0.95$, $K_{max} = 10$, $K_{min} = 2$, $n = 15$
		\STATE Set $MaxGeneration = 250$, $num\_runs = 6$, $initial\_prob = 0.2$, $final\_prob = 0.1$
		\STATE Load dataset from "MyData.txt"
		\FOR{each run from 1 to $num\_runs$}
		\STATE Call \texttt{findClusters} to find the best firefly and fitness
		\STATE Store best solution and fitness
		\ENDFOR
		\STATE Identify best solution based on lowest fitness		
	\end{algorithmic}
\end{algorithm}
\FloatBarrier
\begin{algorithm}[!h]
	\footnotesize
	\caption{findClusters($\alpha$, $\gamma$, $\delta$, $K_{max}$, $K_{min}$, $n$, $data\_set$, $MaxGeneration$, $initial\_prob$, $final\_prob$)}
	\label{al:algorithm3}
	\begin{algorithmic}[1]
		\STATE Generate initial fireflies with random centroids between $K_{min}$ and $K_{max}$
		\FOR{each iteration from 1 to $MaxGeneration$}
		\STATE Calculate adaptive add/remove probability based on iteration
		\STATE Compute fitness for all fireflies and Sort fireflies by fitness
		\STATE Update best firefly if current best is better
		\STATE Move fireflies towards better ones using \texttt{move\_firefly}
		\STATE Adjust centroids with add/remove probability
		\STATE Update randomness coefficient $\alpha$
		\ENDFOR
		\STATE Return the best firefly and its fitness
	\end{algorithmic}
\end{algorithm}
\FloatBarrier
\begin{algorithm}[!h]
	\footnotesize
	\caption{Calculate\_fitness(unique\_fireflies, $n$, $data\_set$)}
	\label{al:algorithm4}
	\begin{algorithmic}[1]
		\FOR{each firefly}
		\STATE Assign points to nearest centroids to form clusters
		\STATE Calculate compactness as average distance between points and centroids
		\STATE Calculate separation as average distance between centroids
		\STATE Calculate total TSP penalty using a heuristic for the traveling salesman problem
		\STATE Combine compactness, separation, and TSP penalty to compute the fitness score
		\ENDFOR
		\STATE Return the fitness scores
	\end{algorithmic}
\end{algorithm}
\FloatBarrier
\begin{algorithm}[!h]
	\footnotesize
	\caption{move\_firefly(unique\_fireflies, Lightn, current\_prob, $n$, $\gamma$, $\alpha$, $K_{min}$, $K_{max}$)}
	\label{al:algorithm5}
	\begin{algorithmic}[1]
		\FOR{each firefly}
		\FOR{each other firefly with better fitness}
		\STATE Move firefly towards better firefly by adjusting centroids
		\STATE Probabilistically adjust number of centroids
		\ENDFOR
		\ENDFOR
		\STATE Return moved fireflies
	\end{algorithmic}
\end{algorithm}
\FloatBarrier

\subsubsection{Movement of fireflies}

When two fireflies are of equal length (K), then there will be no issue of moving the less brighter firefly towards the brighter one (\textbf{use equation \ref{eqn:move}}). 
In automatic clustering different fireflies have different K values and hence have different lengths. To align with that, we introduced a movement strategy of \textbf{finding the nearest centroids in Firefly 2 and moving each centroid of Firefly 1 toward that nearest centroid} using the following formula:
\begin{equation}
	\begin{split}
		\text{new centroid position\_FF1} = & \ (1 - \beta) \times \text{Centroid of Firefly 1} \\
		& + \beta \times \text{Centroid of Firefly 2} \\
		& + \alpha \times \text{randomness}
	\end{split}
	\label{moveFunc}
\end{equation}


\subsubsection{Why the Fitness Function Works Well?}

The fitness function consists of three main components: compactness, separation, and the Traveling Salesman Problem (TSP) penalty. Each component addresses a specific aspect of clustering quality.

\noindent
Compactness measures how closely the data points within a cluster are grouped around their centroid. It is calculated as the average distance of all the points in a cluster from the centroid. In clustering, minimizing the distance between points within a cluster is critical because it ensures that the cluster members are highly similar to one another. Lower compactness values indicate tighter and more cohesive clusters. Therefore, the algorithm encourages the formation of well-defined clusters with minimal intra-cluster variance. The inclusion of compactness in the fitness function helps generate clusters that are meaningful, interpretable, and internally consistent.

\noindent
Separation refers to the average distance between the centroids of different clusters. High separation implies that clusters are well-distinguished from one another, reducing overlap and ambiguity between cluster boundaries. This property is important because it ensures that data points belonging to different clusters remain sufficiently distinct. By incorporating separation into the fitness function, the algorithm promotes the creation of clusters that are not only internally cohesive but also externally well-separated. This improves the overall discriminative quality of the clustering process, especially when handling datasets with overlapping or closely positioned groups.
	
\noindent
	The TSP penalty evaluates how well-organized the internal structure of each cluster is by using a Traveling Salesman Problem (TSP) heuristic to estimate the total traversal distance among points within a cluster. This component discourages the formation of clusters with irregular, scattered, or poorly structured spatial arrangements. Instead, it encourages clusters whose internal point distributions are spatially coherent and organized. The TSP penalty is particularly valuable in spatial clustering problems where the relative arrangement of points within clusters is important. As a result, this measure improves clustering quality by introducing an additional structural constraint beyond simple distance minimization.
	
\noindent
The combination of compactness, separation, and the TSP penalty creates a multi-objective optimization framework for evaluating clustering quality. Each component captures a different characteristic of an effective clustering solution. Compactness focuses on minimizing intra-cluster distances, separation emphasizes maximizing inter-cluster distances, and the TSP penalty enhances the internal structural organization of clusters. The use of weighting parameters $(\alpha, \beta$, and $\gamma)$ allows the algorithm to balance these objectives according to the requirements of the problem. For example, assigning a larger weight to compactness prioritizes tighter clusters, whereas increasing the weight of separation promotes more distinct cluster boundaries. This flexibility enables the algorithm to adapt effectively to datasets with varying cluster sizes, shapes, and densities.

\noindent
The integration of compactness, separation, and the TSP penalty into a unified fitness function provides a comprehensive assessment of clustering quality. Together, these measures optimize multiple aspects of clustering simultaneously, including internal cohesion, external distinctiveness, and spatial organization. This multi-objective perspective results in more balanced and robust clustering solutions. Furthermore, the adjustable weighting mechanism provides flexibility, enabling the algorithm to adapt to different application scenarios and dataset characteristics. Such adaptability is particularly useful when dealing with complex datasets containing clusters with diverse structures and distributions.

\subsubsection{Why the Movement Strategy Works Well?}

\noindent
The movement strategy in the Firefly Algorithm is designed to efficiently explore the search space while gradually improving clustering quality. In the attraction-based movement mechanism, fireflies with higher fitness values, representing poorer clustering solutions, move toward fireflies with lower fitness values, representing better solutions. The degree of attraction depends on the distance between fireflies, where nearby fireflies exert stronger influence than distant ones. This mechanism acts as a local search strategy that guides the population toward promising regions of the search space. By continuously adjusting centroid positions based on interactions with better-performing fireflies, the algorithm incrementally refines clustering quality over successive iterations. This strategy effectively balances exploitation of high-quality solutions with broader exploration of the search space, helping the algorithm avoid poor local minima while still converging toward improved clustering configurations.\\

\noindent
In addition to attraction toward better solutions, the movement equation incorporates a randomness component controlled by the parameter $\alpha$. Even when a firefly moves toward another firefly with superior fitness, random perturbations are introduced into the movement process. The inclusion of randomness is important because it prevents the population from converging too rapidly to a local optimum. Without sufficient randomness, the search process may become trapped in suboptimal regions of the search space. By introducing controlled random exploration, the algorithm is able to investigate new and potentially better regions that may otherwise remain unexplored. This mechanism helps maintain diversity among the fireflies and improves the algorithm’s ability to escape local optima, ultimately increasing the likelihood of identifying globally optimal clustering solutions.\\

\noindent
A key feature of the proposed approach is its ability to dynamically adapt the number of clusters K during the optimization process. Unlike traditional clustering algorithms such as KKK-means, where the number of clusters must be specified beforehand, the Firefly Algorithm allows fireflies with different KKK values to interact with one another. Through these interactions, centroids may be probabilistically added or removed based on the relative proximity and quality of neighboring fireflies. This adaptive mechanism enables the algorithm to automatically discover an appropriate number of clusters that best represents the underlying structure of the dataset. The flexibility to modify KKK dynamically is particularly beneficial for complex datasets where the true number of clusters is unknown in advance. As a result, the algorithm becomes more robust and data-driven, producing clustering solutions that better capture natural data distributions.

\paragraph{Why It Works Well for Clustering}

\begin{enumerate}
	\item Exploration and Exploitation\\
	
	The firefly algorithm strikes an effective balance between exploration (randomness and firefly movement) and exploitation (attraction to better solutions). This is crucial for solving complex clustering problems where the search space is high-dimensional or multi-modal.
	\item Dynamic Adaptation\\
	
	The algorithm dynamically adapts the number of clusters, enabling it to find the optimal \( K \) without predefining the number of clusters. This makes it more flexible and versatile compared to fixed-cluster algorithms like \( K \)-means.
\end{enumerate}

\noindent
The firefly algorithm's fitness function, which combines compactness, separation, and the TSP penalty, provides a robust evaluation of cluster quality. The movement strategy, which includes attraction toward better solutions and randomness, ensures that the algorithm can efficiently explore the search space while avoiding local optima. Together, these strategies make the firefly algorithm highly effective for clustering problems, providing both flexibility and adaptability in discovering the optimal clusters.

	\section{Experimentation and Results}
This chapter shows the experimental results of the simulations carried out for solving clustering problem using the firefly algorithm. Section \ref{S1}, shows data sets used for the study as requested by the robotic navigation problem. Results obtained for fixed clusters are presented in Section \ref{S3}. Finally results obtained for automatic clustering with FA are demonstrated in Section \ref{S4}

\subsection{Data Sets}\label{S1}
\begin{figure}[!h]
	\centering
	
	\begin{minipage}{0.48\textwidth}
		\centering
		\includegraphics[width=0.45\textwidth]{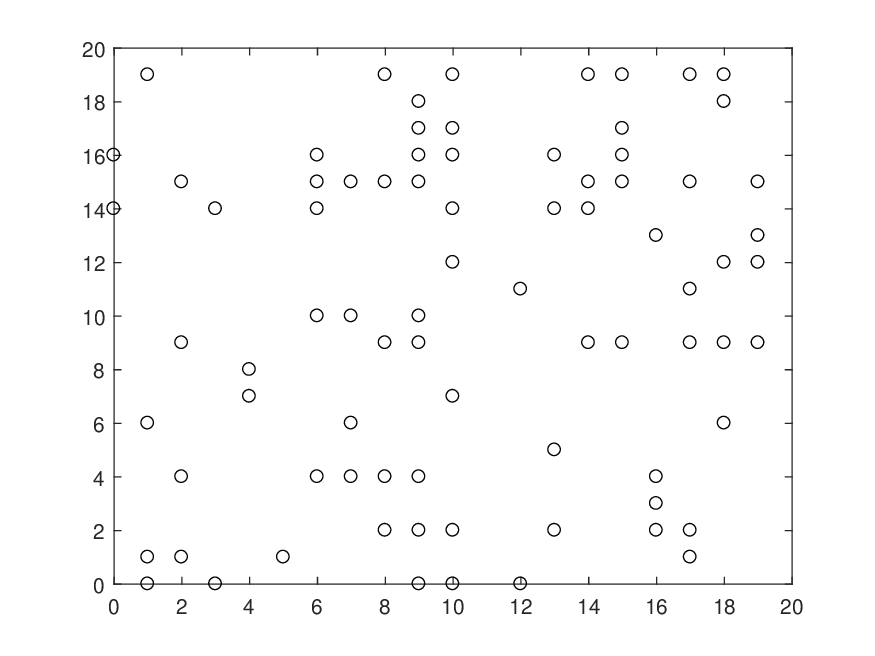}
		\caption{Data Set 1 - 80 points}
		\label{fig:DS1}
	\end{minipage}
	\hfill
	\begin{minipage}{0.48\textwidth}
		\centering
		\includegraphics[width=0.6\textwidth]{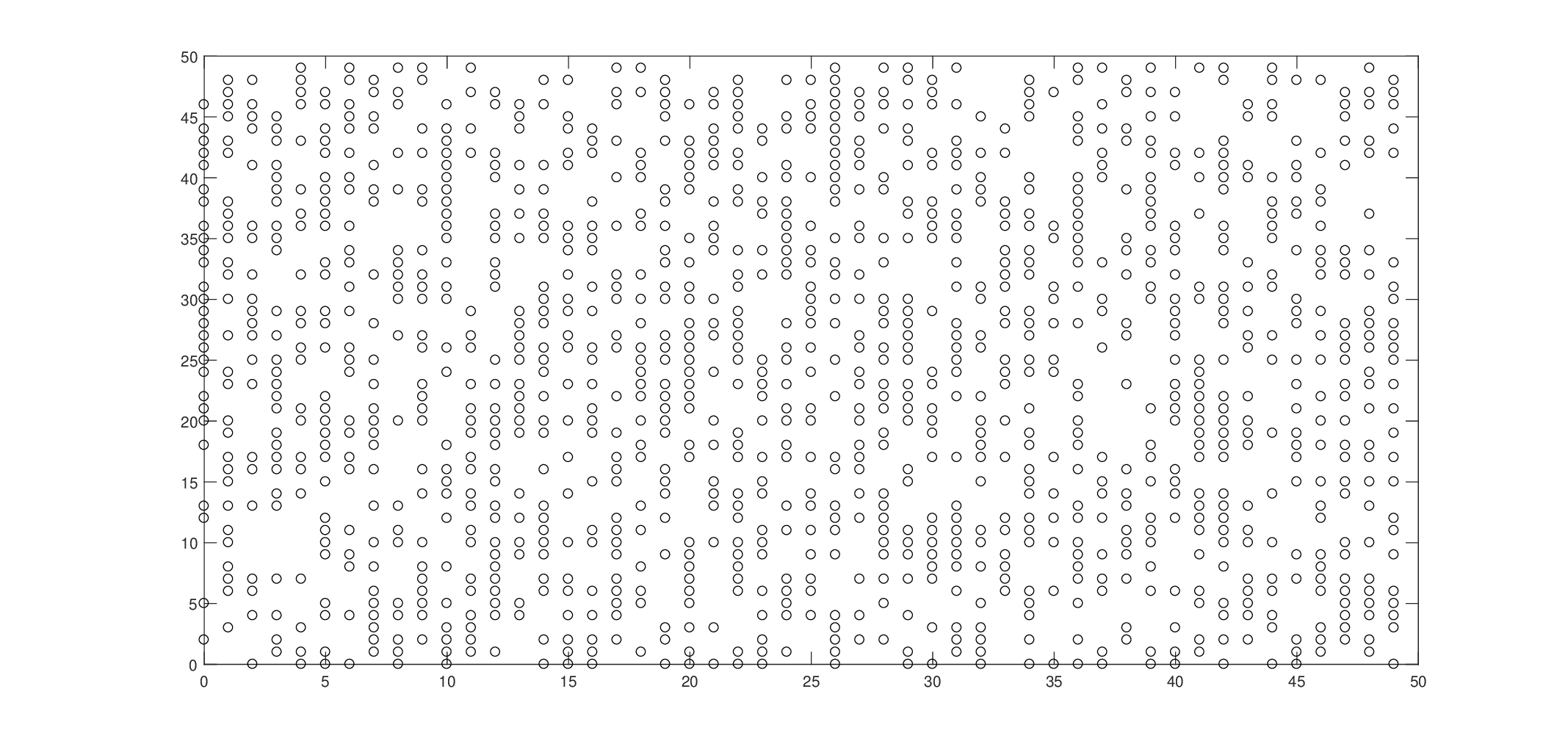}
		\caption{Data Set 2 - 1250 points}
		\label{fig:DS22}
	\end{minipage}
	
\end{figure}

The focus of this research is on the deployment of a
robotic sensor network to persistently and effectively monitor multiple locations spread across a large field, which is 2D. Therefore the study focused on clustering data on 2D spaces. The results were obtained mainly for two fields having 80 coordinates and 1250 coordinates. The spread of the data in given in figure \ref{fig:DS1} and figure \ref{fig:DS22}.
\FloatBarrier
\FloatBarrier

In this research, we used convex hulls for the boundary representation of the found clusters.

\subsection{Parameters used in the study}
The parameters and their values used in the study is given in \texttt{Table \ref{para}}.

\FloatBarrier
\begin{table}[!h]
	\footnotesize
		\caption{Parameters and their values}
	\label{para}
	\begin{tabular}{|ll|ll|}
		\hline
		\multicolumn{2}{|l|}{}                                      & \multicolumn{2}{c|}{\textbf{Values}}                                                   \\ \hline
		\multicolumn{1}{|l|}{}                & \textbf{Parameters} & \multicolumn{1}{l|}{\textbf{Data Set 1}}                         & \textbf{Data Set 2} \\ \hline
		\multicolumn{1}{|l|}{Clustering with FA for known K values}   & Fireflies           & \multicolumn{1}{l|}{15 - 25}                                     & 15 - 25             \\ \hline
		\multicolumn{1}{|l|}{}                                        & Alpha               & \multicolumn{1}{l|}{0.3}                                         & 0.3                 \\ \hline
		\multicolumn{1}{|l|}{}                                        & Gamma               & \multicolumn{1}{l|}{1}                                           & 1                   \\ \hline
		\multicolumn{1}{|l|}{}                                        & Delta               & \multicolumn{1}{l|}{0.95}                                        & 0.95                \\ \hline
		\multicolumn{1}{|l|}{}                                        & Iterations          & \multicolumn{1}{l|}{100}                                         & 100                 \\ \hline
		
		\multicolumn{1}{|l|}{}                & \textbf{Parameters} & \multicolumn{1}{l|}{\textbf{Data Set 1}} & \textbf{Data Set 2} \\ \hline
		\multicolumn{1}{|l|}{Clustering with FA for unknown K values} & Fireflies           & \multicolumn{1}{l|}{15 - 25}                                     & 15 - 25             \\ \hline
		\multicolumn{1}{|l|}{}                                        & Alpha               & \multicolumn{1}{l|}{0.5}                                         & 0.5                 \\ \hline
		\multicolumn{1}{|l|}{}                                        & Gamma               & \multicolumn{1}{l|}{1}                                           & 1                   \\ \hline
		\multicolumn{1}{|l|}{}                                        & Delta               & \multicolumn{1}{l|}{0.95}                                        & 0.95                \\ \hline
		\multicolumn{1}{|l|}{}                                        & Iterations          & \multicolumn{1}{l|}{250}                                         & 250                 \\ \hline
		\multicolumn{1}{|l|}{}                                        & K                   & \multicolumn{1}{l|}{3-10}                                        & 3-10                \\ \hline
	\end{tabular}
\end{table}
\FloatBarrier

\subsection{Clustering with FA for known K values}\label{S3}

This is a similar approach to K means clustering. However, use of meta-heuristics like firefly algorithms can mitigate some of the disadvantages, K means adhere, such as 
sensitive to initial centroids, limited effectiveness for complex shapes, equal-size cluster assumption, computationally inefficient for large datasets, and, assumes continuous features. These issues can be well handled by the firefly algorithm.\\

We evaluated WCSS - within-cluster sum of distances (squared) (or inertia) here is the total sum of squared distances from each data point to its respective cluster centroid, over different K values. Results were obtained with both the firefly algorithm and the K means clustering algorithm. Using elbow method, we tried to comapare the results of two algorithms.

\subsubsection{Data Set 1 - Clustering Results}

\FloatBarrier
\begin{figure}[!h]
	\centering
	
	\begin{minipage}{0.48\textwidth}
		\centering
		\tiny
		\begin{tabular}{|l|l|}
			\hline
			\textbf{K} & \textbf{WCSS} \\ \hline
			2  & 3147.4 \\ \hline
			3  & 2230.4 \\ \hline
			4  & 1523.2 \\ \hline
			5  & 1296.8 \\ \hline
			6  & 945.6  \\ \hline
			7  & 803.3  \\ \hline
			8  & 847.4  \\ \hline
			9  & 618.3  \\ \hline
			10 & 631.7  \\ \hline
		\end{tabular}
		\captionof{table}{Within-cluster sum of squares (WCSS) for Dataset 1 using the firefly algorithm}
	\end{minipage}
	\hfill
	\begin{minipage}{0.48\textwidth}
		\centering
		\includegraphics[width=0.6\textwidth]{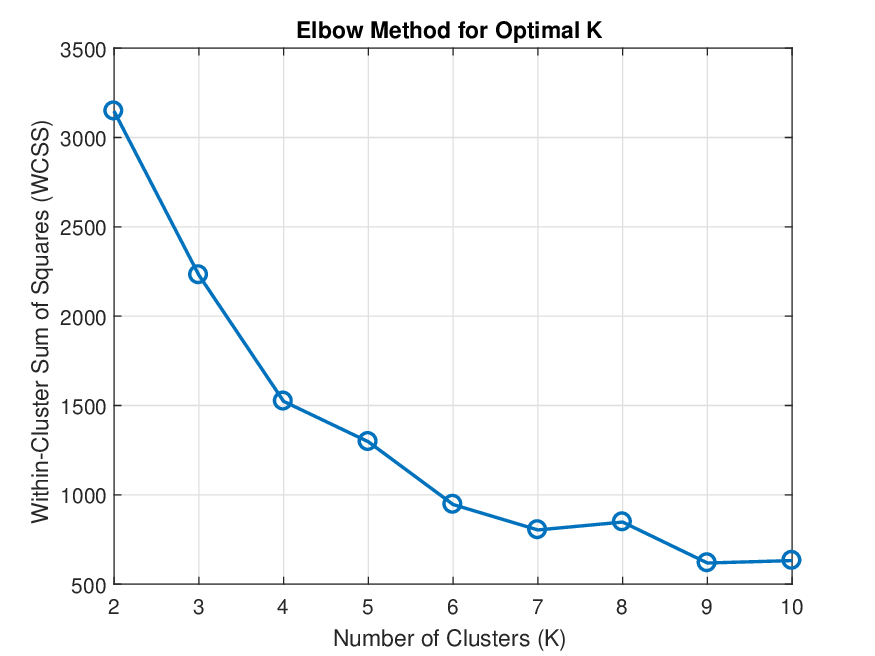}
		\caption{WCSS vs $K$ for Dataset 1 using the firefly algorithm}
		\label{fig:DS222}
	\end{minipage}
\end{figure}
\FloatBarrier
\FloatBarrier
\vspace*{-0.5cm}
The WCSS values show a sharp decrease up to \(K = 4\) or \(5\), followed by diminishing improvements, indicating the elbow point. Thus, \(K = 4\) or \(5\) is a suitable choice, balancing clustering quality and complexity. For comparison, K-means was also applied to the same dataset to obtain corresponding WCSS values.

The WCSS shows a sharp decrease up to \(K = 4\) or \(5\), after which improvements diminish, indicating the elbow point. Both firefly and K-means algorithms produce similar cost values and suggest the same optimal \(K\). Thus, the firefly algorithm achieves comparable performance to K-means while addressing its limitations.
\FloatBarrier
\begin{figure}[!h]
	\centering
	
	\begin{minipage}{0.48\textwidth}
		\centering
		\tiny
		\begin{tabular}{|l|l|}
			\hline
			\textbf{K} & \textbf{WCSS} \\ \hline
			2  & 2954.8798 \\ \hline
			3  & 1863.6966 \\ \hline
			4  & 1303.6024 \\ \hline
			5  & 1105.3929 \\ \hline
			6  & 913.3595  \\ \hline
			7  & 749.3099  \\ \hline
			8  & 566.6984  \\ \hline
			9  & 452.9060  \\ \hline
			10 & 445.1397  \\ \hline
		\end{tabular}
		\captionof{table}{Within-cluster sum of squares (WCSS) for Dataset 1 using the K-Means algorithm}
	\end{minipage}
	\hfill
	\begin{minipage}{0.48\textwidth}
		\centering
		\includegraphics[width=0.6\textwidth]{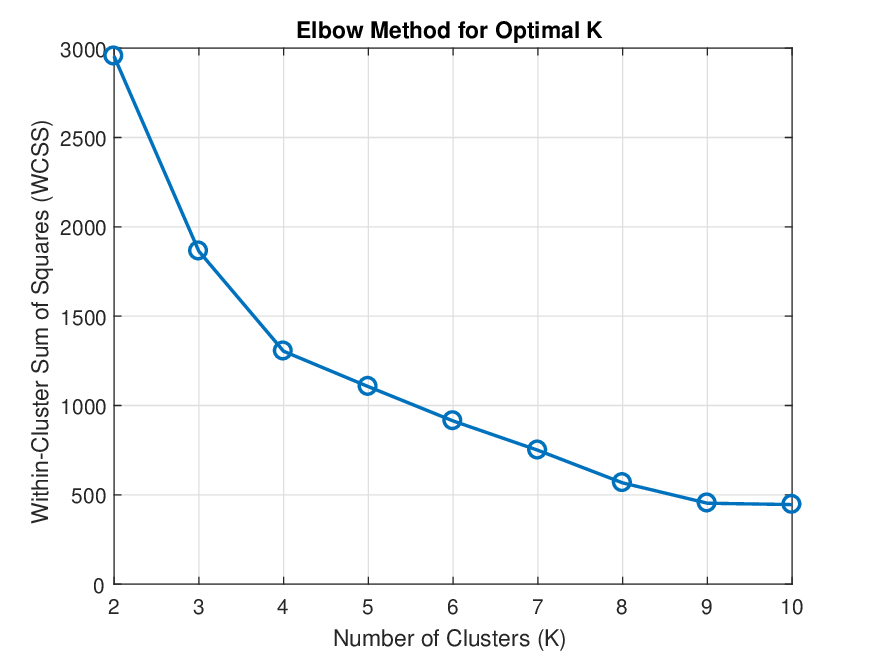}
		\caption{WCSS vs $K$ for Dataset 1 using the K-Means algorithm}
		\label{fig:DS2222}
	\end{minipage}
\end{figure}
\FloatBarrier

\subsubsection{Data Set 2 - Clustering Results}
The analysis of the WCSS values reveals a clear elbow pattern, with a significant reduction observed between K=3 and K=5, indicating substantial improvements in cluster compactness within this range. Beyond K=5, the rate of decrease in WCSS becomes more gradual, reflecting diminishing returns as additional clusters contribute less to improving clustering quality. The elbow point is therefore identified around K=5 or K=6, suggesting that this range provides an optimal balance between compactness and model simplicity. Consequently, selecting K=5 or K=6 is most appropriate, as further increases offer minimal benefit. A similar trend is observed for Dataset 2 with 1250 points, where both the firefly algorithm and K-means produce consistent results and indicate comparable optimal values of K, demonstrating agreement between the two approaches.
\FloatBarrier\FloatBarrier
\begin{figure}[!h]
	\centering
	
	\begin{minipage}{0.48\textwidth}
		\centering
		\tiny
		\begin{tabular}{|l|l|}
			\hline
			\textbf{K} & \textbf{WCSS} \\ \hline
			3  & 3.0247e+05 \\ \hline
			4  & 1.5697e+05 \\ \hline
			5  & 1.2527e+05 \\ \hline
			6  & 1.0495e+05 \\ \hline
			7  & 8.2046e+04 \\ \hline
			8  & 7.0352e+04 \\ \hline
			9  & 6.7068e+04 \\ \hline
			10 & 5.8100e+04 \\ \hline
		\end{tabular}
		\captionof{table}{Within-cluster sum of squares (WCSS) for Dataset 2 using the firefly algorithm}
	\end{minipage}
	\hfill
	\begin{minipage}{0.48\textwidth}
		\centering
		\includegraphics[width=0.4\textwidth]{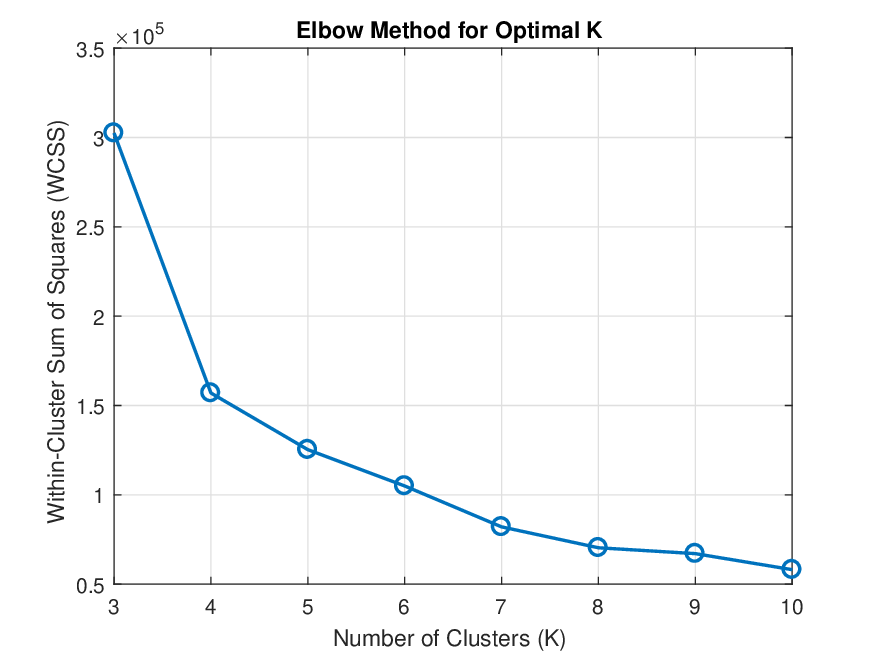}
		\caption{WCSS vs $K$ for Dataset 2 using the firefly algorithm}
		\label{fig:DS4}
	\end{minipage}
	
\end{figure}
\FloatBarrier
\vspace*{-0.5cm}
\FloatBarrier
\begin{figure}[!h]
	\centering
	
	\begin{minipage}{0.48\textwidth}
		\centering
		\tiny
		\begin{tabular}{|l|l|}
			\hline
			\textbf{K} & \textbf{WCSS} \\ \hline
			3  & 205011.38 \\ \hline
			4  & 131953.87 \\ \hline
			5  & 112755.11 \\ \hline
			6  & 96119.98  \\ \hline
			7  & 77234.89  \\ \hline
			8  & 66015.42  \\ \hline
			9  & 56563.22  \\ \hline
			10 & 52691.09  \\ \hline
		\end{tabular}
		\captionof{table}{Within-cluster sum of squares (WCSS) for Dataset 2 using the K-Means algorithm}
	\end{minipage}
	\hfill
	\begin{minipage}{0.48\textwidth}
		\centering
		\includegraphics[width=0.4\textwidth]{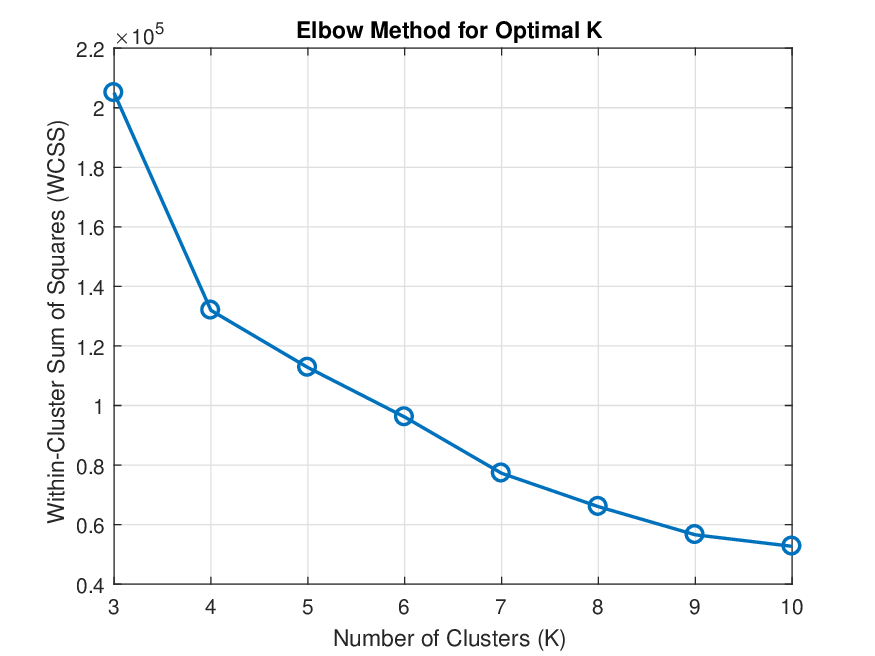}
		\caption{WCSS vs $K$ for Dataset 2 using the K-Means algorithm}
		\label{fig:DS4_kmeans}
	\end{minipage}
	
\end{figure}
\FloatBarrier

\noindent
\textbf{Visualizing clusters with convex hulls}

We adhere the convex hulls to visually present the boundaries of clusters. Figures \ref{fig:figure1} and \ref{fig:figure2} show convex hulls obtained for the optimal K values for both data sets using the firefly algorithm.
\FloatBarrier
\begin{figure}[!h]
	\centering
	\begin{minipage}{0.45\textwidth}
		\centering
		\includegraphics[width=\textwidth]{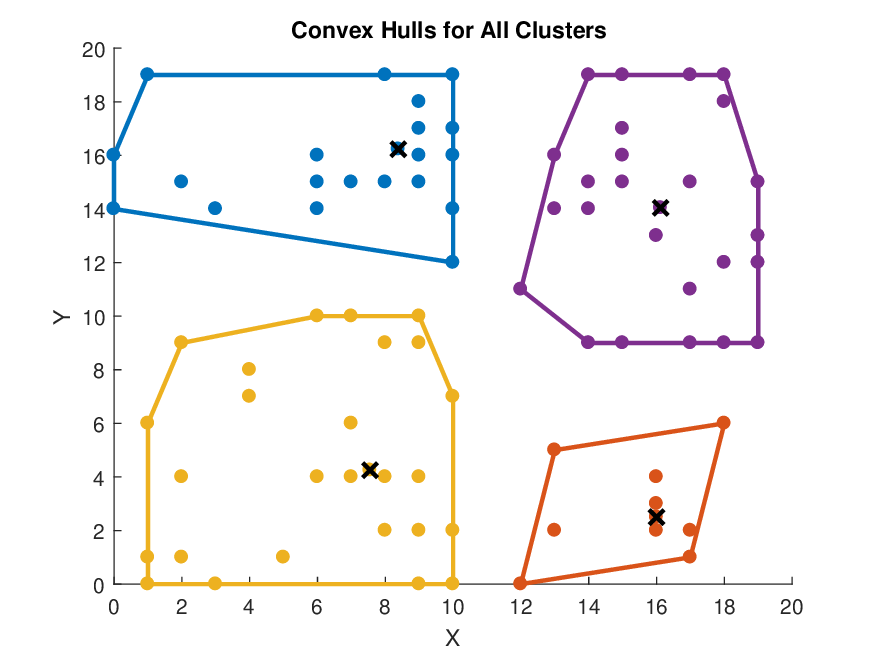}
		\caption{K=4 Centroids obtained with FF - Data set 1}
		\label{fig:figure1}
	\end{minipage}\hfill
	\begin{minipage}{0.45\textwidth}
		\centering
		\includegraphics[width=\textwidth]{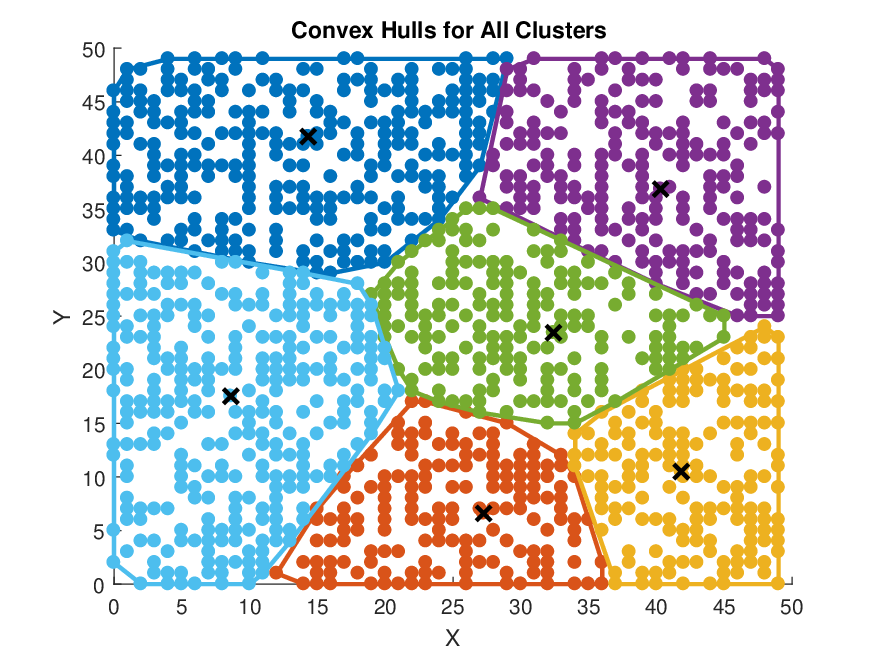}
		\caption{K=6 Centroids obtained with FF - Data set 2}
		\label{fig:figure2}
	\end{minipage}
	\caption{Convex hulls for the optimal K values for both data sets using the firefly algorithm}
\end{figure}
\FloatBarrier

\subsection{Automatic Clustering with FA}\label{S4}

Clustering with fixed K is an achievable task, where clustering for unknown K is challenging. Thus lot of research were carried out for such an approach. But in most of such research, they have run the algorithm several times to obtain the most trusted K value. In our approach, to the original firefly algorithm, we introduced new fitness function and centroid movement strategy rather than moving the whole fireflies towards the brighter ones to achieve the objective of the study as well as to automatically find the most suitable K value.

\subsubsection{Data set 1 - Clustering Results}
Here we considered the new fitness function, where we considered compactness, separation and TSP penalty, to find better clusters for the research question of effective robot navigation. 
To obtain precise solutions, we run the algorithm 6 times and plot the K value given at each run and the fitness at the given K. As the final result we selected the solution (centroids) and the K with best fitness.

For all the solutions in all runs, we evaluate the WCSS values as well. It is not for the elbow approach, but to see whether there are significant fluctuations in the WCSS values. Since all the 6 runs gives good results, there should not be much variations in the WCSS values.
Obtained centroids are
(3, 7), (7, 16), (14, 10), (15, 13)

\paragraph{Fitness and WCSS Analysis}

The fitness values across the six runs show low variation, indicating stable performance of the automatic clustering approach. Among these runs, the solution corresponding to \(K = 4\) appears most frequently and yields the lowest fitness values, suggesting it as the most suitable number of clusters identified by the algorithm.

The WCSS values (\(6453, 6482, 9030, 8190, 5857, 7053\)) also demonstrate moderate variation across runs, further confirming the consistency of the clustering results. Overall, both fitness and WCSS analyses indicate that the algorithm reliably converges to similar-quality solutions, with \(K = 4\) emerging as the preferred cluster configuration.
\FloatBarrier
\begin{figure}[!h]
	\centering
	
	\begin{minipage}{0.48\textwidth}
		\centering
		\includegraphics[width=0.6\textwidth]{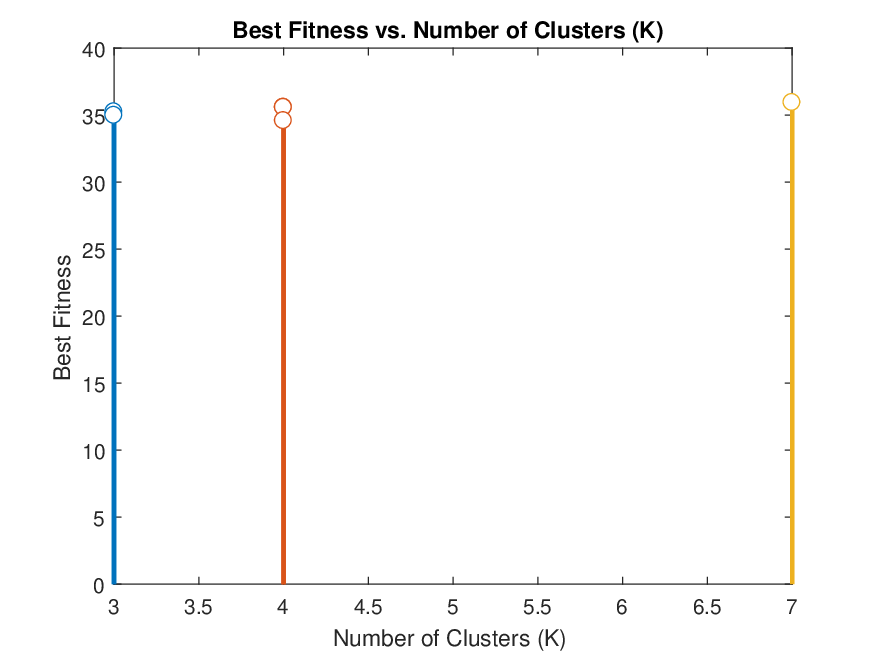}
		\caption{Fitness vs $K$ for Dataset 1 (Automatic clustering with FA)}
		\label{fig:DS55}
	\end{minipage}
	\hfill
	\begin{minipage}{0.48\textwidth}
		\centering
		\includegraphics[width=0.6\textwidth]{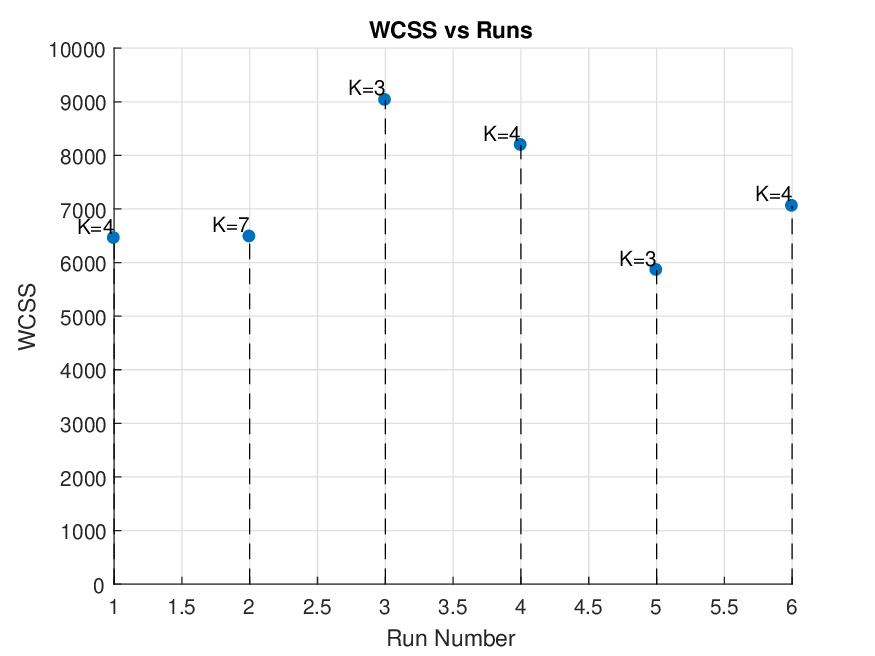}
		\caption{WCSS vs runs (Automatic clustering with FA)}
		\label{fig:DS77}
	\end{minipage}
	
\end{figure}
\FloatBarrier
\subsubsection{Data set 2 - Clustering Results}
\FloatBarrier
\begin{figure}[!h]
	\centering
	\includegraphics[width=0.5\textwidth]{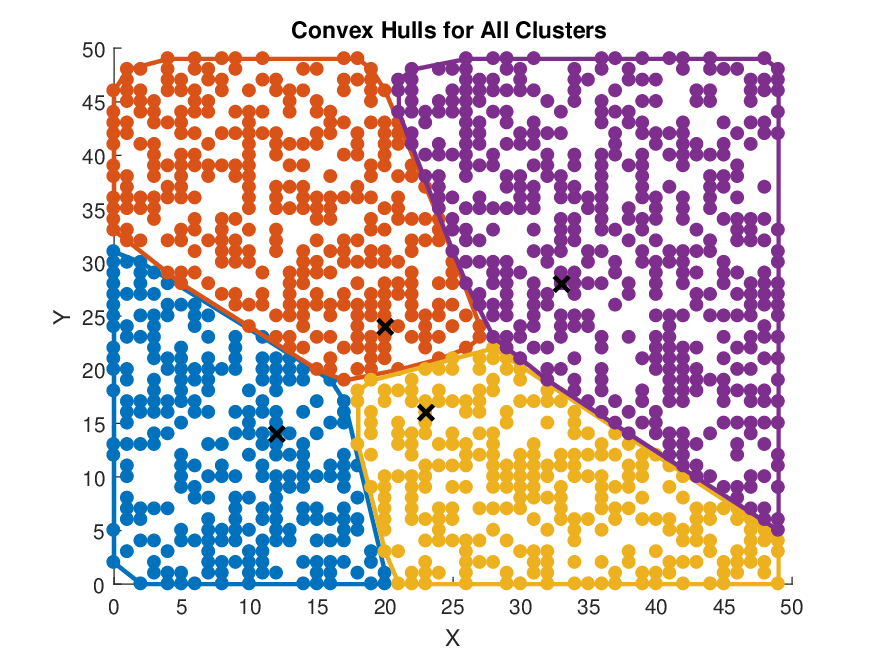}
	\caption{Final Clusters for the Data set 2 - Automatic clustering with FA}
	\label{fig:DS45}
\end{figure}
\FloatBarrier
\noindent
Obtained centroids are (12    14), (20    24), (23    16), (33    28)\\

\paragraph{Fitness and WCSS Across Runs}

The fitness values across the six runs demonstrate relatively low variation, indicating stable performance of the automatic clustering process. This suggests that the algorithm consistently converges to solutions of similar quality.

The corresponding WCSS values (\(928535, 1668041, 713439, 575330, 1153769, 864266\)) exhibit noticeable fluctuations across runs. However, despite this variation, the values remain within a reasonable range, indicating that all runs produce acceptable clustering solutions. These fluctuations can be attributed to the stochastic nature of the algorithm, but they do not significantly affect the overall clustering quality.

Overall, both fitness and WCSS analyses confirm that the algorithm maintains robustness and consistency across multiple runs while still allowing some diversity in the generated solutions.
\FloatBarrier
\begin{figure}[!h]
	\centering
	
	\begin{minipage}{0.48\textwidth}
		\centering
		\includegraphics[width=0.6\textwidth]{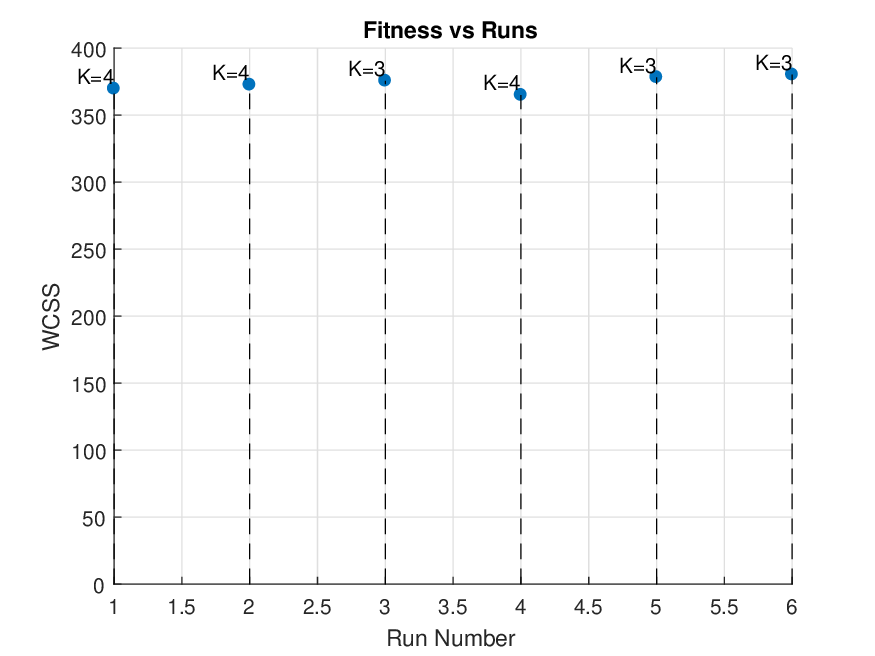}
		\caption{Fitness vs runs (Automatic clustering with FA)}
		\label{fig:DS6}
	\end{minipage}
	\hfill
	\begin{minipage}{0.48\textwidth}
		\centering
		\includegraphics[width=0.6\textwidth]{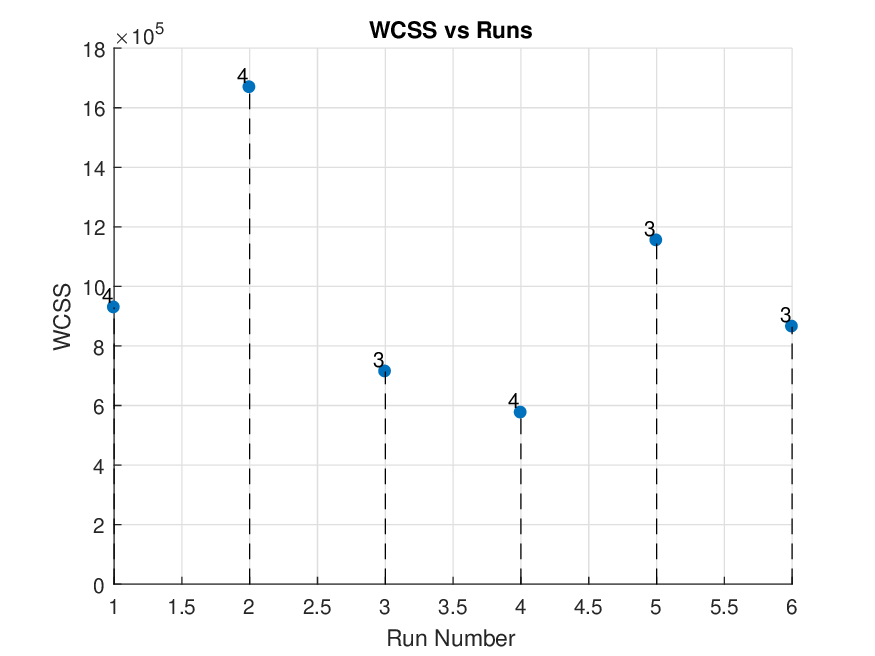}
		\caption{WCSS vs runs (Automatic clustering with FA)}
		\label{fig:DS7}
	\end{minipage}
	
\end{figure}
\FloatBarrier
General Trend: While there is a spread, the values do not exhibit wild, unpredictable swings; they are all in a similar magnitude range (hundreds of thousands). This suggests that while the exact values differ, they might represent a consistent pattern or trend when taken as a whole.

Cluster Interpretation: In clustering, WCSS measures within-cluster variability, so having some fluctuation across runs isn't unusual. The fluctuations here might reflect minor adjustments due to random initialization or slight differences in clustering but not necessarily a large variance that would invalidate the results.

Practical Consistency: For practical purposes, these values are relatively close when we consider that clustering algorithms often show some variation across runs, especially with initializations or data sensitivity. The values are still comparable, suggesting that the clustering structure is stable across these runs.

Conclusion: Although there is variation, the values aren’t drastically different, and they can be interpreted as reasonably close in the context of clustering, where some variability is expected. These fluctuations suggest a stable clustering outcome across runs rather than major inconsistencies.
\subsection{Comparing with K means}
To compare the new algorithm with K means, whether the objectives of the research is accomplished, we evaluated the optimized routes for the clusters for both algorithms. Ant Colony Systems (ACS) algorithm was applied to the clusters obtained by both automatic firefly algorithm and the K means. The total path distance is evaluated.

\FloatBarrier
\begin{figure}[!h]
	\centering
	\begin{minipage}{0.45\textwidth}
		\centering
		\includegraphics[width=0.9\textwidth]{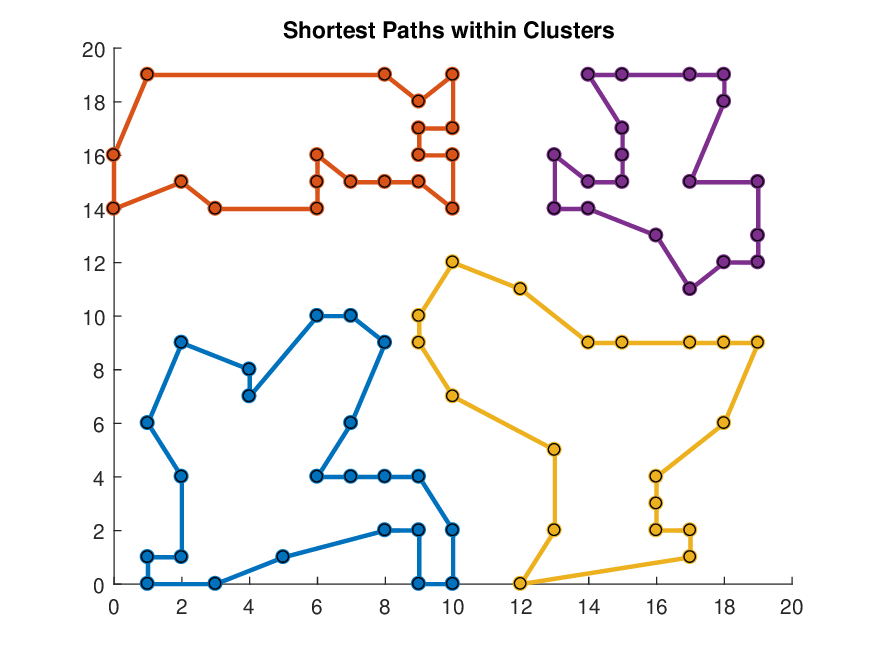}
		\caption{Shortest Paths for the Clusters with FA}
		\label{fig:figure11}
	\end{minipage}\hfill
	\begin{minipage}{0.45\textwidth}
		\centering
		\includegraphics[width=0.9\textwidth]{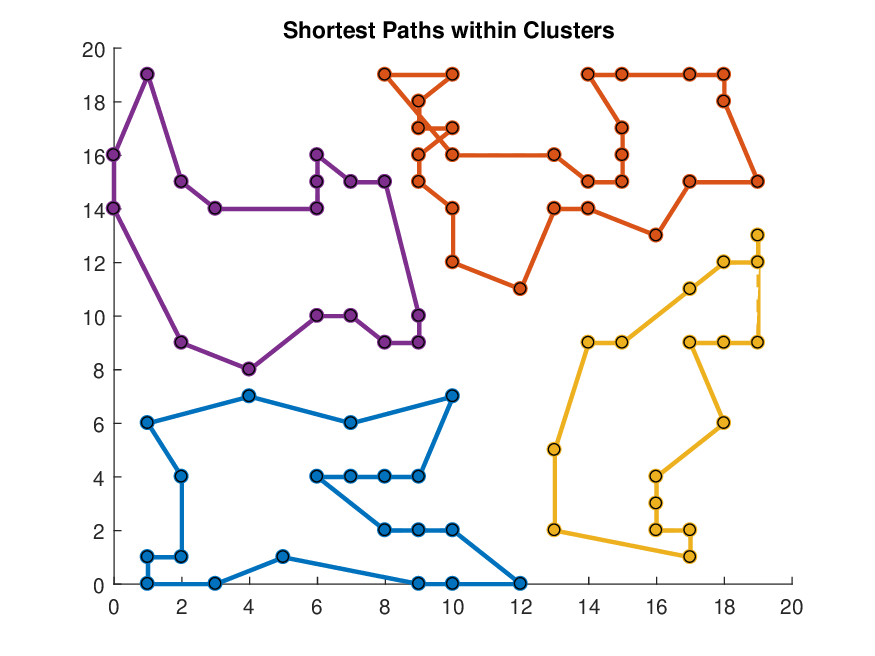}
		\caption{Shortest Paths for the Clusters with K-Means}
		\label{fig:figure22}
	\end{minipage}
	\caption{Shortest Paths for the Clusters with ACS algorithm - Data Se 1}
\end{figure}
\FloatBarrier

\begin{itemize}
	\item Total distance for all 4 clusters by FA =  43.69+35.47+39.47+29.70$\approx$ 148.33
	\item Total distance for all 4 clusters by K-Means= 41.90+45.53+34.48+38.08 $\approx$ 159.99
	
\end{itemize}

\FloatBarrier
\begin{figure}[!h]
	\centering
	\begin{minipage}{0.45\textwidth}
		\centering
		\includegraphics[width=\textwidth]{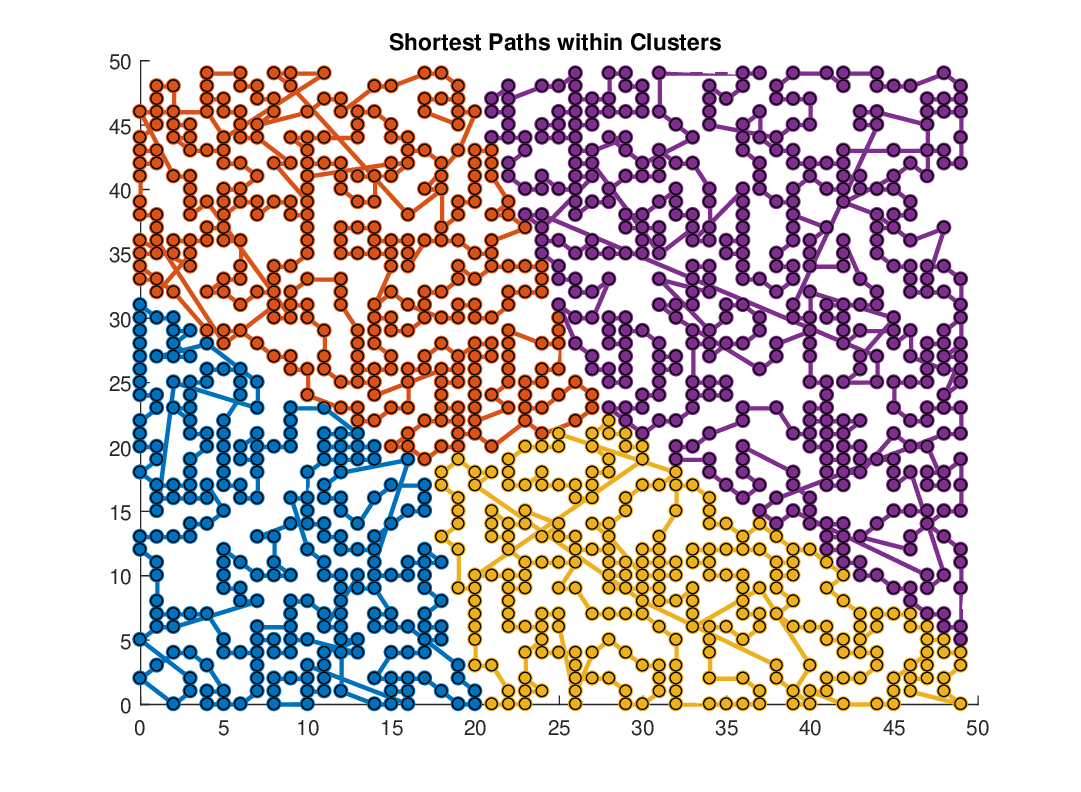}
		\caption{Shortest Paths for the Clusters with FA}
		\label{fig:figure1}
	\end{minipage}\hfill
	\begin{minipage}{0.45\textwidth}
		\centering
		\includegraphics[width=\textwidth]{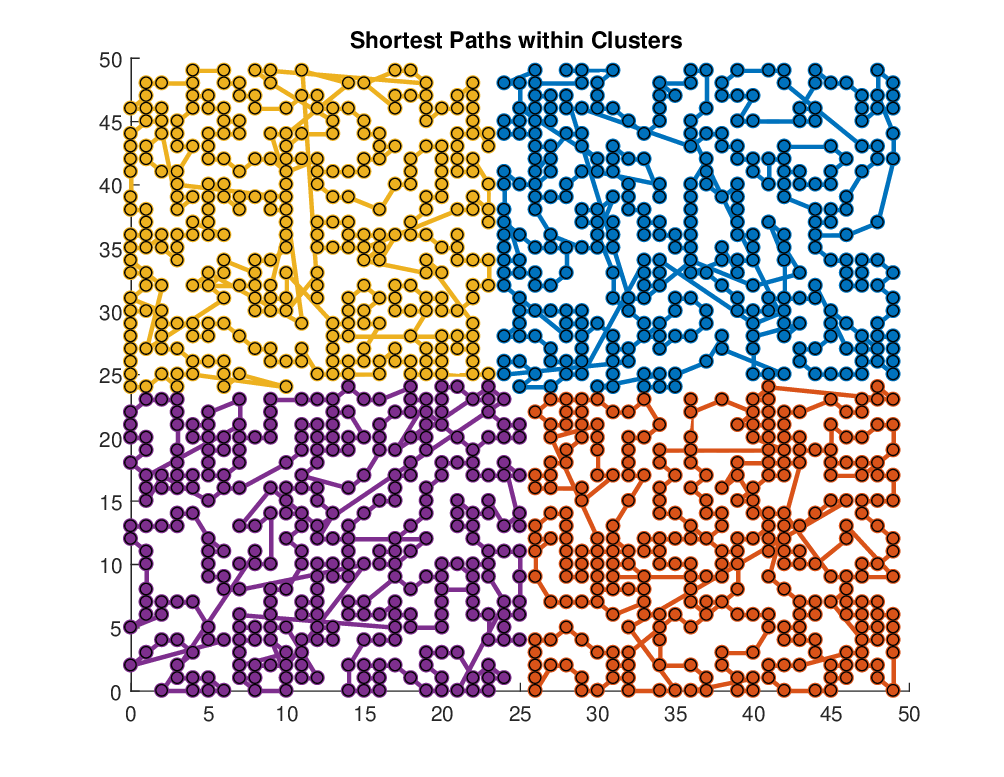}
		\caption{Shortest Paths for the Clusters with K-Means}
		\label{fig:figure2}
	\end{minipage}
	\caption{Shortest Paths for the Clusters with ACS algorithm - Data Se 2}
\end{figure}
\FloatBarrier

\begin{itemize}
	\item Total distance for all 4 clusters by FA = 333.63+456.81+361.18+ 643.14 $\approx$ 1793
	\item Total distance for all 4 clusters by K-Means
	= 519.46+446.15+480.16+486.93 $\approx$ 1931
	
\end{itemize}

It indicate that K-Means gives clusters based on the distance to the centers, it is not successful in providing shortest routes for navigation. On the other-hand, FA, automatically gives cluster centers and still good in providing shorter paths compared to the K means algorithm.

\section{Conclusion and Future Work}
This study aimed to develop a clustering approach that is independent of the predefined number of clusters (\(K\)) while ensuring efficient within-cluster navigation for the given monitoring system. To address this, a modified firefly algorithm incorporating a novel fitness function and a centroid-based movement strategy was proposed. The results demonstrate that the proposed meta-heuristic approach is capable of effectively identifying suitable cluster structures without prior specification of \(K\), while maintaining good clustering quality.

The findings highlight that the performance of clustering algorithms is highly dependent on the characteristics of the dataset, and that the choice of clustering method should be guided not only by data proximity but also by the underlying objectives of the application. Furthermore, given the NP-hard nature of clustering, meta-heuristic techniques such as the firefly algorithm provide a flexible and powerful alternative to traditional methods. The proposed modification of the firefly algorithm proves particularly useful in scenarios where initialization sensitivity must be avoided, the number of clusters needs to be determined automatically, and clustering objectives extend beyond simple distance-based criteria.

For future work, several directions can be explored to enhance the proposed approach. First, the firefly algorithm involves several algorithm-specific parameters that currently require manual tuning; developing a self-adaptive or self-tuning mechanism would improve its usability and robustness. Second, extending the investigation to include other meta-heuristic algorithms could provide comparative insights and potentially yield further improvements. Finally, the current study is limited to two-dimensional datasets, and evaluating the proposed method on higher-dimensional data would help assess its scalability and general applicability.

	\bibliographystyle{vancouver}
	\bibliography{references}  
	
\end{document}